\documentclass[preprint,4pt]{elsarticle}

\usepackage{tikz}
\usetikzlibrary{positioning,arrows.meta, quotes}
\usepackage[bookmarks=false]{hyperref}
    \hypersetup{colorlinks,
      linkcolor=blue,
      citecolor=blue,
      urlcolor=blue}

\usepackage{amsmath}
\usepackage{amssymb}
\usepackage{enumitem}
\usepackage{graphicx}
\usepackage{multirow}
\usepackage{adjustbox}
\usepackage{array}
\usepackage{tabularx}
\usepackage{geometry}

\usepackage{amssymb}

\journal{}

\begin{document}

\begin{frontmatter}





\title{Causal Time-Series Synchronization for Multi-Dimensional Forecasting}

\author[a]{Michael Mayr}

\affiliation[a]{organization={Software Competence Center Hagenberg},
            addressline={Softwarepark 32a}, 
            city={Hagenberg},
            postcode={4232}, 
            country={Austria}}

\author[a]{Georgios C. Chasparis}
\author[b]{Josef Küng}

\affiliation[b]{organization={Institute for Application-Oriented Knowledge Processing, Johannes Kepler University},
            addressline={Altenberger Str. 69}, 
            city={Linz},
            postcode={4040}, 
            country={Austria}}


\begin{abstract}
The process industry's high expectations for Digital Twins require modeling approaches that can generalize across tasks and diverse domains with potentially different data dimensions and distributional shifts i.e., Foundational Models. Despite success in natural language processing and computer vision, transfer learning with (self-) supervised signals for pre-training general-purpose models is largely unexplored in the context of Digital Twins in the process industry due to challenges posed by multi-dimensional time-series data, lagged cause-effect dependencies, complex causal structures, and varying number of (exogenous) variables. 
We propose a novel channel-dependent pre-training strategy that leverages synchronized cause-effect pairs to overcome these challenges by breaking down the multi-dimensional time-series data into pairs of cause-effect variables. Our approach focuses on: (i) identifying highly lagged causal relationships using data-driven methods, (ii) synchronizing cause-effect pairs to generate training samples for channel-dependent pre-training, and (iii) evaluating the effectiveness of this approach in channel-dependent forecasting. Our experimental results demonstrate significant improvements in forecasting accuracy and generalization capability compared to traditional training methods.
\end{abstract}

\end{frontmatter}




\section{Introduction}
\label{main}

The number of sensors and the corresponding data produced in the process industry continuously increases, enabling a rich source of sensor data \cite{Lasi2014}, which presents an unprecedented opportunity to harness complex sensor data for enhancing operational efficiency and decision-making. Manufacturers in the European Union, in particular, face the dual challenge of accelerating decision-making processes while conforming with ambitious energy-efficiency standards to reduce greenhouse gas emissions from production plants \cite{EU2050}. These challenges underscore the need for intelligently managed and controlled manufacturing lines, highlighting Digital Twins as the key enabler to increase competitiveness, sustainability, and efficiency \cite{Kritzinger2018}. A Digital Twin (DT) in the process industry represents a virtual model of a production system, integrating sensor data, simulations, and real-time data processing to predict and optimize the production process \cite{Rasheed2020}. Recently, the evolution of DTs embraces cognitive capabilities, i.e. cognitive DTs (see \cite{Abburu2020}, \cite{Mayr2024}), showcasing their progression towards autonomy and intelligence. A cognitive DT should be capable of learning from and providing decision-support for various tasks and use cases, including classification and regression-related tasks, necessitating the training across various datasets with different multi-dimensional data characteristics. Transfer Learning (TL) and Self-Supervised Learning (SSL) are promising research directions and may tackle the complex task of developing such "higher-level" DTs \cite{Mayr2024}. \\

\tikzset{
    block/.style = {draw, fill=green!20, rectangle, minimum height=4.5em, minimum width=4.5em},
    arrow/.style={-Stealth, thick},
}
\tikzset{every edge quotes/.style =
          { fill = green!20,
            sloped,
            execute at begin node = $,
            execute at end node   = $  }}
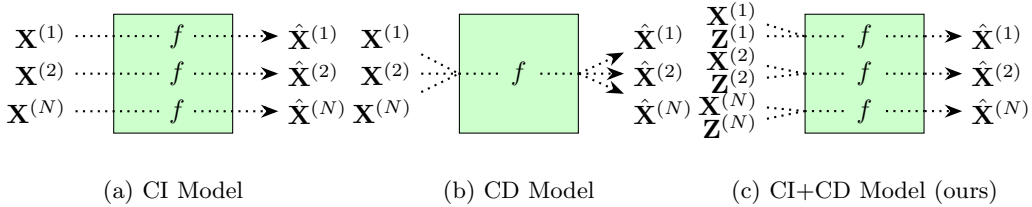
\begin{figure}[h!]
\centering
\begin{tikzpicture}[node distance=2cm, >=Stealth]

    \node [block] (forecaster1) {};
    \node (input1) [left=0.5cm of forecaster1, yshift=0.5cm] {$\mathbf{X}^{(1)}$};
    \node (input2) [left=0.5cm of forecaster1] {$\mathbf{X}^{(2)}$};
    \node (input3) [left=0.5cm of forecaster1, yshift=-0.5cm] {$\mathbf{X}^{(N)}$};
    \node (output1) [right=0.6cm of forecaster1, yshift=0.5cm] {$\hat{\mathbf{X}}^{(1)}$};
    \node (output2) [right=0.6cm of forecaster1] {$\hat{\mathbf{X}}^{(2)}$};
    \node (output3) [right=0.6cm of forecaster1, yshift=-0.5cm] {$\hat{\mathbf{X}}^{(N)}$};
    
    \draw [arrow, dotted] (input1) to ["f"] (output1);
    \draw [arrow, dotted] (input2) to ["f"] (output2);
    \draw [arrow, dotted] (input3) to ["f"] (output3);
    \node at (forecaster1.south) [below=0.5cm of forecaster1] {\small (a) CI Model};

    \node [block] (forecaster2) [right=3cm of forecaster1] {};
    \node (input2a) [left=0.5cm of forecaster2, yshift=0.5cm] {$\mathbf{X}^{(1)}$};
    \node (input2b) [left=0.5cm of forecaster2, yshift=0cm] {$\mathbf{X}^{(2)}$};
    \node (input2c) [left=0.5cm of forecaster2, yshift=-0.5cm] {$\mathbf{X}^{(N)}$};

    \node (output2a) [right=0.6cm of forecaster2, yshift=0.5cm] {$\hat{\mathbf{X}}^{(1)}$};
    \node (output2b) [right=0.6cm of forecaster2, yshift=0cm] {$\hat{\mathbf{X}}^{(2)}$};
    \node (output2c) [right=0.6cm of forecaster2, yshift=-0.5cm] {$\hat{\mathbf{X}}^{(N)}$};

    \draw [thick,dotted] (input2a) -- (forecaster2.west);
    \draw [thick,dotted] (input2b) -- (forecaster2.west);
    \draw [thick,dotted] (input2c) -- (forecaster2.west);
    
    \draw [thick, dotted] (forecaster2.west) to ["f"] (forecaster2.east);

    \draw [arrow, dotted] (forecaster2.east) -- (output2a);
    \draw [arrow, dotted] (forecaster2.east) -- (output2b);
    \draw [arrow, dotted] (forecaster2.east) -- (output2c);
    
    \node at (forecaster2.south) [below=0.5cm of forecaster2] {\small (b) CD Model};

    \node [block] (forecaster3) [right=3cm of forecaster2] {};
    \node (input3a) [left=0.5cm of forecaster3, yshift=0.8cm] {$\mathbf{X}^{(1)}$};
    \node (input3b) [left=0.5cm of forecaster3, yshift=0.5cm] {$\mathbf{Z}^{(1)}$};
    \node (input4a) [left=0.5cm of forecaster3, yshift=0.2cm] {$\mathbf{X}^{(2)}$};
    \node (input4b) [left=0.5cm of forecaster3, yshift=-0.1cm] {$\mathbf{Z}^{(2)}$};
    \node (input5a) [left=0.5cm of forecaster3, yshift=-0.4cm] {$\mathbf{X}^{(N)}$};
    \node (input5b) [left=0.5cm of forecaster3, yshift=-0.7cm] {$\mathbf{Z}^{(N)}$};

    \node (output3a) [right=0.5cm of forecaster3, yshift=0.5cm] {$\hat{\mathbf{X}}^{(1)}$};
    \node (output4a) [right=0.5cm of forecaster3, yshift=0cm] {$\hat{\mathbf{X}}^{(2)}$};
    \node (output5a) [right=0.5cm of forecaster3, yshift=-0.5cm] {$\hat{\mathbf{X}}^{(N)}$};
    
    \draw [thick, dotted] (input3a) -- ([shift={(0,-0.3)}] forecaster3.north west);
    \draw [thick, dotted] (input3b) -- ([shift={(0,-0.3)}] forecaster3.north west);
    \draw [thick, dotted] (input4a) -- (forecaster3.west);
    \draw [thick, dotted] (input4b) -- (forecaster3.west);
    \draw [thick, dotted] (input5a) -- ([shift={(0,0.3)}] forecaster3.south west);
    \draw [thick, dotted] (input5b) -- ([shift={(0,0.3)}] forecaster3.south west);
    
    \draw [arrow, dotted] ([shift={(0,-0.3)}]  forecaster3.north east) -- (output3a);
    \draw [arrow, dotted] (forecaster3.east) -- (output4a);
    \draw [arrow, dotted] ([shift={(0,0.3)}] forecaster3.south east) -- (output5a);
    
    \draw [thick, dotted] (forecaster3.west) to ["f"] (forecaster3.east);
    \draw [thick, dotted] ([shift={(0,-0.3)}] forecaster3.north west) to ["f"] ([shift={(0,-0.3)}] forecaster3.north east);
    \draw [thick, dotted] ([shift={(0,0.3)}] forecaster3.south west) to ["f"]  ([shift={(0,0.3)}] forecaster3.south east);

    \node at (forecaster3.south) [below=0.5cm of forecaster3] {\small (c) CI+CD Model (ours)};
\end{tikzpicture}
\caption{(a) Channel Independence (CI), i.e. each variable is treated as an isolated univariate problem only indirectly learning from other channels through shared weights; (b) Channel Dependence (CD), i.e. all variables are mixed and treated as a multivariate problem directly learning from other variables; (c) Proposed hybrid model (CI+CD), i.e. target variables $\mathbf{X}^{(N)}$ are mixed with exogenous variables $\mathbf{Z}^{(N)}$ forming blocks of cause-effect pairs, directly learning cause-effect behaviour of two variables through CD and indirectly learning from other cause-effects from other variable interactions through shared weights, i.e. CI. Illustrative example adapted from Wang et al. \cite{Wang2024}.}
\label{fig:cdvsci}
\end{figure}

Time-series data in the process industry comprises different causally related variables, and accurate modelling requires capturing the relationships between these. Several channel-dependent (CD) strategies jointly model multiple variables using GNNs (\cite{Yi2023}, \cite{Huang2023}), MLPs (\cite{Ekambaram2023}, \cite{Wang2024}), CNNs (\cite{Wu2023}), Transformers (\cite{Ni2023}, \cite{Liu2024}) \cite{Zhao2024}. Unexpectedly, while only modelling cross-time dependence, recently proposed channel-independent (CI) methods (\cite{Nie2023}, \cite{Chen2024}, \cite{Lee2024}) often outperform CD methods on various benchmarks \cite{Zhao2024}, possibly due to misaligned time-series and noise of non-causally related variables. An illustrative example of CD and CI forecasting is depicted in Fig.\ref{fig:cdvsci}. Training general-purpose models utilizing CD strategies offers clear advantages in the process industry, where variables influence the process state, such as control inputs, raw material properties, and environmental factors. In addition, Digital Twins rely on simulation and control capabilities, necessitating cross-variable dependence on exogenous information when forecasting target variables. Furthermore, the physical counterpart of the Digital Twin in the process industry often exhibits high inertia, meaning changes to a controllable variable may affect the system only after a certain delay, with the system slowly reaching equilibrium. Capturing such complex behaviour necessitates not only a CD approach, it is also important to consider that highly-lagged cross-variable dependency may not be captured by the context window, i.e. the (historical) data the model is conditioned on, of such models.

\subsection{Contributions}
\label{sec:Contributions}
\begin{figure}[h!]
    \centering
\begin{tikzpicture}

\definecolor{lightgray}{gray}{0.9}
\definecolor{midgray}{gray}{0.6}
\definecolor{darkgray}{gray}{0.3}

\def\boxwidth{1}
\def\xstart{0}

\draw[dotted, thick] (-0.35, 3.5) -- (-0.15,3.5);

\draw[fill=white] (\xstart, 3) rectangle ({\xstart+\boxwidth}, 4);
\draw[fill=lightgray] ({\xstart+\boxwidth}, 3) rectangle ({\xstart+2*\boxwidth}, 4);
\draw[fill=midgray] ({\xstart+2*\boxwidth}, 3) rectangle ({\xstart+3*\boxwidth}, 4);
\draw[fill=darkgray] ({\xstart+3*\boxwidth}, 3) rectangle ({\xstart+4*\boxwidth}, 4);
\draw[fill=midgray] ({\xstart+4*\boxwidth}, 3) rectangle ({\xstart+5*\boxwidth}, 4);

\draw[dotted, thick] ({\xstart+5*\boxwidth+0.1}, 3.5) -- ({\xstart+5*\boxwidth+0.3},3.5);
\node at (-1.2, 3.5) {\small Effect $\mathbf{X}^{(i)}$};

\draw[dotted, thick] (-0.35, 2) -- (-0.15,2);

\draw[fill=midgray] (\xstart, 1.5) rectangle ({\xstart+\boxwidth}, 2.5);
\draw[fill=darkgray] ({\xstart+\boxwidth}, 1.5) rectangle ({\xstart+2*\boxwidth}, 2.5);
\draw[fill=midgray] ({\xstart+2*\boxwidth}, 1.5) rectangle ({\xstart+3*\boxwidth}, 2.5);
\draw[fill=lightgray] ({\xstart+3*\boxwidth}, 1.5) rectangle ({\xstart+4*\boxwidth}, 2.5);
\draw[fill=white] ({\xstart+4*\boxwidth}, 1.5) rectangle ({\xstart+5*\boxwidth}, 2.5);

\draw[dotted, thick] ({\xstart+5*\boxwidth+0.1}, 2) -- ({\xstart+5*\boxwidth+0.3},2);
\node at (-1.2, 2) {\small Cause $\mathbf{Z}^{(j)}$};

\draw[->] (\xstart, 1) -- ({\xstart+5*\boxwidth}, 1) node[midway, below] {\small Time $t$};
\node at ({\xstart+5*\boxwidth / 2}, 0) {\small (a) non-synchronized cause-effect pair};
\def\boxwidth{1}
\def\xstart{6}

\draw[dotted, thick] (-0.35, 3.5) -- (-0.15,3.5);

\draw[fill=white] (\xstart, 3) rectangle ({\xstart+\boxwidth}, 4);
\draw[fill=lightgray] ({\xstart+\boxwidth}, 3) rectangle ({\xstart+2*\boxwidth}, 4);
\draw[fill=midgray] ({\xstart+2*\boxwidth}, 3) rectangle ({\xstart+3*\boxwidth}, 4);
\draw[fill=darkgray] ({\xstart+3*\boxwidth}, 3) rectangle ({\xstart+4*\boxwidth}, 4);
\draw[fill=midgray] ({\xstart+4*\boxwidth}, 3) rectangle ({\xstart+5*\boxwidth}, 4);

\draw[dotted, thick] ({\xstart+5*\boxwidth+0.1}, 3.5) -- ({\xstart+5*\boxwidth+0.3},3.5);

\draw[dotted, thick] (-0.35, 2) -- (-0.15,2);

\draw[fill=white] (\xstart, 1.5) rectangle ({\xstart+\boxwidth}, 2.5);
\draw[fill=lightgray] ({\xstart+\boxwidth}, 1.5) rectangle ({\xstart+2*\boxwidth}, 2.5);
\draw[fill=midgray] ({\xstart+2*\boxwidth}, 1.5) rectangle ({\xstart+3*\boxwidth}, 2.5);
\draw[fill=darkgray] ({\xstart+3*\boxwidth}, 1.5) rectangle ({\xstart+4*\boxwidth}, 2.5);
\draw[fill=midgray] ({\xstart+4*\boxwidth}, 1.5) rectangle ({\xstart+5*\boxwidth}, 2.5);
\draw[arrow] (\xstart, 2) -- ({\xstart+2*\boxwidth}, 2) node[midway, below] {\small ($t+\delta_{ij})$};

\draw[dotted, thick] ({\xstart+5*\boxwidth+0.1}, 2) -- ({\xstart+5*\boxwidth+0.3},2);

\draw[->] (\xstart, 1) -- ({\xstart+5*\boxwidth}, 1) node[midway, below] {\small Time $t$};
\node at ({\xstart+5*\boxwidth / 2}, 0) {\small
(b) synchronized cause-effect pair};

\end{tikzpicture}
\caption{Illustrative example of a non-synchronized cause-effect pair (a), the effect lags with a time delay of two time-steps and a synchronized cause-effect pair (b), where the cause is shifted $\delta_{ij}$ time-steps, i.e. a shift by two time-steps in this example, to account for the causal lag. Illustrative example adapted from Zhao et al.~\cite{Zhao2024}.}
\label{fig:CauseEffectPair}
\end{figure}
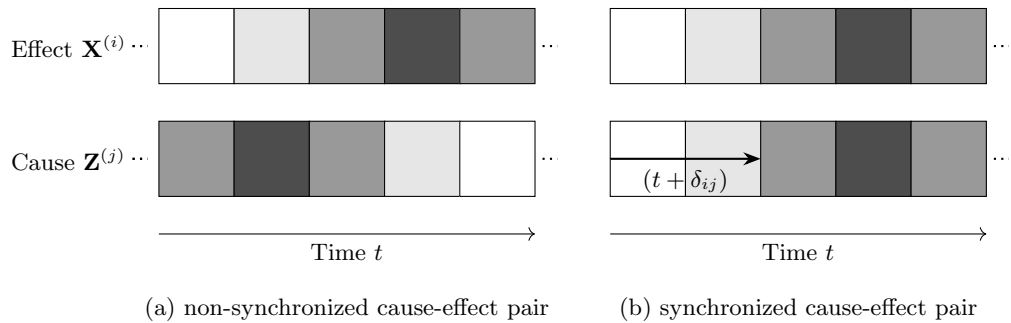

In this paper, we tackle the challenges of applying CD pre-training on multi-dimensional time-series datasets with complex causal structures or isolated causal clusters in addition to slow inertia and time misalignment, characteristics commonly encountered in the process industry. These characteristics may make it difficult for a CD pre-training strategy to efficiently converge across multi-dimensional datasets, as the models may constantly fit potential non-causally related variables and time-mismatched cause-and-effect behaviours. A related time-alignment problem is also encountered in a recently proposed control-flow DT for the process industry \cite{Mayr2022}; however, it is not tackled in their paper.
We propose a novel hybrid pre-training strategy (see Fig.~\ref{fig:cdvsci} (c)) based on synchronized cause-effect pairs (see Fig.~\ref{fig:CauseEffectPair}) to mitigate the mentioned problems and to focus pre-training specifically on cause-effect behaviours, an essential part of simulation and control in industrial processes, by breaking down complex causal structures into smaller, more focused training samples. 

This contribution underscores the urgent need for more robust, specialized pre-training strategies to handle such complex data effectively and addresses the following research questions: 

\begin{enumerate}[label=\textbf{Research Question \arabic*:}, leftmargin=*, itemsep=0.5em]
    \item \textit{How can highly lagged relationships in industrial processes be identified and modelled using data-driven approaches?}
    \item \textit{How can one synchronize cause-effect pairs w.r.t causal lag and create context-horizon training samples for CD pre-training without losing information in the sample?} 
    \item \textit{Do CD pre-training methods benefit from cause-effect synchronization for regression-related tasks like forecasting?}
\end{enumerate}

This research has several sections that address the above-stated research questions systematically. Sec.~\ref{sec:background} discusses predictive modelling and Digital Twins in the context of the process industry. We highlight the problem of process inertia and complex causal structures for generalized process modelling. Sec.~\ref{sec:methodology} explains extracting a causal relationship graph using linear Granger Causality. Additionally, we propose to break up complex causal structures of multi-dimensional time-series data into pairs of synchronized cause-effect training samples. In Sec.~\ref{sec:results}, we validate the proposed methodology on synthetically generated data with known causal relationships on the task of forecasting. In Sec.~\ref{sec:conclusion}, we synthesize the findings from the conducted research and the experimental results, evaluate the stated research questions and discuss shortcomings and possible solutions.

\section{Background and Related Work}
\label{sec:background}

\subsection{Predictive Modelling in Digital Twins for Process Industry}
Time-series forecasting, i.e. extrapolation of time-series into the future, a commonly encountered regression problem for Digital Twins in process industry \cite{Mayr2024}, is a critical area of research in process industry and a necessity for intelligent decision support systems. Forecasting problems are often tackled statistically in the first step, e.g., ARIMA, VAR, or Exponential Smoothing, commonly acting as strong baselines for more advanced models. Advanced deep sequential modelling methodologies for regression-related tasks in the process industry are recurrent neural networks (RNNs), long-short-term memory networks (LSTMs) and CNNs. Mayr et al. \cite{Mayr2024} provide a detailed structured literature review of the latest modelling methodologies and learning paradigms for DTs in process industries. 
Some of the state-of-the-art deep learning models have shown promising results for generalization capabilities across various domains (i.e. different datasets) and tasks (i.e. regression-related and classification-related), which enabled researchers to pre-train time-series foundation models on time-series big data gathered from various domains. (e.g. TimeGPT \cite{Garza2024}, TinyTimeMixer \cite{EkambaramTiny2024}, Lag-Llama \cite{Rasul2024}). Due to the different number of dimensions in the various datasets, many foundation models are pre-trained using CI approaches (e.g. \cite{EkambaramTiny2024}, \cite{Rasul2024}). In contrast, the latest development for CD approaches include Moirai \cite{Woo2024}, which tackles the different dimensions by a novel adaptation of the attention mechanism, or TinyTimeMixer \cite{EkambaramTiny2024} which merges both CD and CI approaches, keeping the flexibility of CI in pre-training and incorporates CD strategy in the fine-tuning stage in order to natively support multi-dimensional datasets with endogenous and exogenous variables.

\subsection{Data-Driven Granger Causality and Causal Graph Structures}
Granger causality, initially developed for economics \cite{Granger1969}, is increasingly used in various fields to identify cause-effect relationships (e.g. \cite{Ammann2024}). In the following paper, we utilize a similar approach as Ammann et al. \cite{Ammann2024} to identify cause-effect relationships between pairs of stochastic process variables and employ a simple linear VAR model, i.e. auto-regressive, assuming stationarity and linearly model-able cause-effect relationships. For a detailed analysis of the utilized method, we kindly refer to the authors \cite{Ammann2024}. We estimate the full VAR model on the training data and test for a significant increase in variance for specific time lags for pairs of variables using F-tests to denote causality. As a more straightforward use-case, and also for demonstration purposes, we assume globally stationary causal relationships. However, in a real-world scenario, dynamic causal relationships are expected. Zhao et al. \cite{Zhao2024} propose a simple method to directly incorporate dynamic cross-correlation measures, i.e., lead-lags, into the learning procedure. \\
Causal and correlation-related information may be encoded in graph structures, where nodes denote (lagged) variables and edges denote the type of relation, i.e. correlation value, Granger Causality p-value. Encoding such data in graph structures enables complex query capabilities using graph-related query languages (e.g. SPARQL\footnote{\url{https://www.w3.org/TR/sparql11-query/}} for RDF graph data formats, Cypher for Neo4J\footnote{\url{https://neo4j.com/}}, or AQL for ArangoDB\footnote{\url{https://arangodb.com/}}). Embedding such information in graph-like format may not be needed for simple use-cases like retrieving sub-graphs, identifying causally related pairs, i.e. cause-effect pairs, for synchronization (described in Sec.~\ref{subsec:subgraphextraction} and Sec.~\ref{subsec:leadlagsync}), however, will be helpful in case of more complex information embedding in the graph. In this research, we utilize RDF triplets to store this information in a Knowledge Graph (KG) format, i.e. the embedded data is queryable by using SPARQL\footnote{\url{https://www.w3.org/TR/rdf-sparql-query/}} standard, however as the graph grows, one may also utilize optimized databases for storing KGs or graph-related structures like Neo4J or ArangoDB, which all provide tailored graph-query languages like Cypher or AQL. \\

\subsection{Channel Dependence vs. Channel Indpendence}
Time-series data in the process industry comprises different causally related variables, and accurate modelling requires capturing the relationships between these. CI methods just model cross-time dependence (e.g. \cite{Nie2023}, \cite{Chen2024}, \cite{Lee2024}), only indirectly modelling cross-variable dependence through shared weights, whereas CD methods (e.g. \cite{Yi2023}, \cite{Huang2023}, \cite{Ekambaram2023}, \cite{Wang2024}) model cross-time and cross-variable dependence. Both CI and CD methods are commonly window-based, i.e., the forecast is conditioned on N time-points of historic context values. Increasing the context requires retraining the model, and increasing the context length is also computationally costly. This poses a problem in the process industry, as there may be causes and effects between different variables of the multi-dimensional dataset that are highly lagged, e.g. over several hours or even days, e.g. a kiln with high inertia to control inputs, which the specific context size may overlook, i.e. missing out on crucial data-driven supervisory signals for the model to generalize. Additionally, CD modelling approaches using the whole dataset, i.e. mixing all variables, may learn from causally unrelated variables or unsynchronized cause-effect behaviours.  \\
In the following, we propose a CD pre-training methodology for multi-dimensional datasets with highly lagged causal data that can be used to pre-train various deep-learning models. Our proposed methodology aims to break down the problem of CD multi-dimensional time-series modelling by splitting the multi-dimensional data into causally related cause-effect pairs that are time synchronized. We hypothesize that by applying such pre-training methodology for CD methods, we move towards better generalization in case of modelling different multi-dimensional datasets as we break down the very complex problem into smaller, potentially easier solvable sub-problems for the model by only modelling time-synchronized cause-effect pairs. In comparison to the work done by Zhao et al.\cite{Zhao2024}, where they account for dynamic lags using synchronization at each time step, we specifically investigate the capability of performance improvements of CD methods for causal dependencies that may lie outside of the specific context window used. We expect an improvement in accuracy for regression-related tasks due to reduced input and output complexity, higher amounts of data and reduced "noise" of wrongly lagged variables.

\section{Methodology - Synchronized Cause-Effect Pairs for Channel Dependent Pre-Training on Multi-Dimensional Data}
\label{sec:methodology}
This section explains the proposed causal lag synchronization (see Sec.~\ref{subsec:leadlagsync}) as well as the construction of cause-effect data samples (see Sec.~\ref{subsec:dataaugmentation}) in detail. Our proposed methodology aims to break down the problem of CD multi-dimensional time-series modelling by splitting the multi-dimensional data into causally related cause-effect pairs that are time synchronized. To do this, we exhaustively search for all possible cause-effect pairs (also transient connectivity), time synchronize the cause variable with the effect variable by applying a right shift operator with the specific lag to the cause variable (getting highly lagged causes into the context window), create data samples where the context holds all historical values for the respective synchronized cause and additionally the non-synchronized cause (in order to not miss any data due to the shift) and effect variable, the target, i.e. the variable that has to be predicted over a horizon, is, in this case, the future effect.

\subsection{Threshold-based Causal Graph Generation}
\label{subsec:subgraphextraction}
Given a multi-dimensional time-series $\mathbf{X}$, we calculate the Granger causality for each possible variable pair up to lag $L_{max}$. We store the causal measurements for each lag and each variable pair. We calculate the "best" lag w.r.t. to the causal measurement, i.e. p-value, and test the causal hypothesis for the selected lag based on an expert-defined or data-driven defined threshold. Depending on the dataset, this results in a sub-graph, where clusters may form or even long causal chains. We exhaustively search for cause-effect pairs and denote the corresponding lag for each pair as $\delta_{ij}$, which is needed for the target-oriented causal shift, explained in the Sec.~\ref{subsec:leadlagsync}. Fig.~\ref{fig:CauseEffectPair} (a) depicts an illustrative example of a non-synchronized cause-effect pair, where the effect follows the cause with a time delay of two time steps.

\subsection{Target-oriented Causal Shift}
\label{subsec:leadlagsync}

Given the identified cause-effect pairs, including the global lags, we adjust the variables accordingly by shifting $\mathbf{Z}^{(j)}$ w.r.t $\mathbf{X}^{(i)}$ by the corresponding calculated lag $\delta_{ij}$ forward in time, such that cause synchronizes with the effect, i.e. the target (see Fig.~\ref{fig:CauseEffectPair} (b)). The pairs reflect the globally stationary cause-effect relationships across the multi-dimensional data and can be utilized for CD pre-training or end-to-end modelling of e.g. regression-related tasks like forecasting (see Sec.~\ref{subsec:dataaugmentation}). The assumption of global static causal lags may not hold in real-world systems. Thus, further refinement by calculating the dynamic lags for each time step or each domain, e.g. different products, different environmental factors, etc., may be a necessity. However, dynamic causal lags are considered a future research direction but are not followed up on in this paper.

\subsection{Predictive Modelling of Cause-Effects using Channel Dependent Deep Learning Models}
\label{subsec:dataaugmentation}
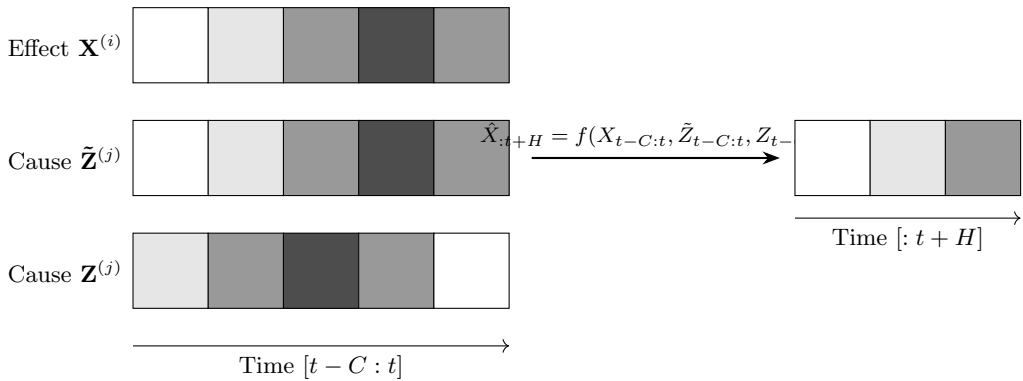
\begin{figure}[ht]
    \centering
\begin{tikzpicture}

\definecolor{lightgray}{gray}{0.9}
\definecolor{midgray}{gray}{0.6}
\definecolor{darkgray}{gray}{0.3}

\def\boxwidth{1}
\def\xstart{-0.3}


\draw[fill=white] (\xstart, 3) rectangle ({\xstart + \boxwidth}, 4);
\draw[fill=lightgray] ({\xstart + \boxwidth}, 3) rectangle ({\xstart + 2*\boxwidth}, 4);
\draw[fill=midgray] ({\xstart + 2*\boxwidth}, 3) rectangle ({\xstart + 3*\boxwidth}, 4);
\draw[fill=darkgray] ({\xstart + 3*\boxwidth}, 3) rectangle ({\xstart + 4*\boxwidth}, 4);
\draw[fill=midgray] ({\xstart + 4*\boxwidth}, 3) rectangle ({\xstart + 5*\boxwidth}, 4);

\node at (-1.2, 3.5) {\small Effect $\mathbf{X}^{(i)}$};


\draw[fill=white] (\xstart, 1.5) rectangle ({\xstart + \boxwidth}, 2.5);
\draw[fill=lightgray] ({\xstart + \boxwidth}, 1.5) rectangle ({\xstart + 2*\boxwidth}, 2.5);
\draw[fill=midgray] ({\xstart + 2*\boxwidth}, 1.5) rectangle ({\xstart + 3*\boxwidth}, 2.5);
\draw[fill=darkgray] ({\xstart + 3*\boxwidth}, 1.5) rectangle ({\xstart + 4*\boxwidth}, 2.5);
\draw[fill=midgray] ({\xstart + 4*\boxwidth}, 1.5) rectangle ({\xstart + 5*\boxwidth}, 2.5);

\node at (-1.2, 2) {\small Cause $\mathbf{\tilde{Z}}^{(j)}$};


\draw[fill=lightgray] (\xstart, 0) rectangle ({\xstart + \boxwidth}, 1);
\draw[fill=midgray] ({\xstart + \boxwidth}, 0) rectangle ({\xstart + 2*\boxwidth}, 1);
\draw[fill=darkgray] ({\xstart + 2*\boxwidth}, 0) rectangle ({\xstart + 3*\boxwidth}, 1);
\draw[fill=midgray] ({\xstart + 3*\boxwidth}, 0) rectangle ({\xstart + 4*\boxwidth}, 1);
\draw[fill=white] ({\xstart + 4*\boxwidth}, 0) rectangle ({\xstart + 5*\boxwidth}, 1);

\node at (-1.2, 0.5) {\small Cause $\mathbf{Z}^{(j)}$};

\draw[->] (\xstart, -0.5) -- ({\xstart + 5*\boxwidth}, -0.5) node[midway, below] {\small Time $[t-C:t$]};

\def\xstart{4.2}

\draw[arrow] ({\xstart + \boxwidth - 0.2}, 2) -- ({\xstart + 4*\boxwidth + 0.1 }, 2) node[midway, above]{\footnotesize $\hat{X}_{:t+H} = f(X_{t-C:t}, \tilde{Z}_{t-C:t}, Z_{t-C:t})$};

\def\xstart{6.5}
\draw[fill=white] (\xstart + 2*\boxwidth, 1.5) rectangle ({\xstart + 3*\boxwidth}, 2.5);
\draw[fill=lightgray] ({\xstart + 3*\boxwidth}, 1.5) rectangle ({\xstart + 4*\boxwidth}, 2.5);
\draw[fill=midgray] ({\xstart + 4*\boxwidth}, 1.5) rectangle ({\xstart + 5*\boxwidth}, 2.5);

\draw[->] (\xstart + 2*\boxwidth, 1.2) -- ({\xstart + 5*\boxwidth}, 1.2) node[midway, below] {\small Time $[:t+H$]};

\end{tikzpicture}
\caption{Illustrative example of the transformed cause-effect pair for channel-dependent forecasting, where the context - $C$ denotes the context length, i.e. historic data points - contains three variables: the historic values of the actual target variable $\mathbf{X}^{(i)}$, the target-shifted, synchronized cause variable $\mathbf{\tilde{Z}}^{(j)}$, and the non-synchronized cause variable $\mathbf{Z}^{(j)}$ to counteract data loss of right shift. The future effect $\hat{X}$ is predicted over a horizon $H$.}

 \label{fig:trainingpairs}
\end{figure}

We construct synchronized subsets of cause-effect training samples using a specific context length and horizon (i.e. prediction horizon). Synchronization leaves data at the beginning of the time series empty (when applied on a sliding window basis, each window would miss some time steps in the beginning). However, in case of causes outside the context window, we would benefit from filling up the missing values by sampling a bigger window to account for potential lags that need to fill up the first shifted missing values. In addition, due to the shift, data is lost in the cause variable as one shifts the latest values outside the context window. This is tackled by specifying a fixed additional cause variable that denotes the non-synchronized version. In this way, no information in the window is lost. Fig.~\ref{fig:trainingpairs} denotes an illustrative example of the transformed cause-effect pair. This input-output structure can be easily used to perform CD pre-training for various state-of-the-art deep learning models. 

\section{Experiments and Results}
\label{sec:results}

\subsection{Experiment 1 - Synchronized vs. Non-Synchronized Cause-Effect Pairs for End-to-End Forecasting}
\label{subsec:exp1}

The following experiment showcases the difference of training with synchronized vs non-synchronized cause-effect pairs with state-of-the-art deep learning models that support exogenous variables, i.e. TSMixerx, MLPMultivariate, BiTCN, and TFT on synthetically generated datasets with known causal lagged relationships. We base the training of the mentioned models on the framework of Olivares et al.\cite{olivares2022}. The training samples are structured as depicted in Fig.~\ref{fig:trainingpairs}, and the models are trained with synchronized cause-effect pairs and non-synchronized cause-effect pairs; we compare the results for regression-related tasks, i.e. in our case, forecasting, in various synthetically generated datasets with known causal relationships at known points in time. We construct the forecasting task with a prediction horizon $H$ of ten time steps and a context length $C$ of three times the horizon, i.e. $C=H*3$ for the experiments. We create ten synthetic datasets of 5 variables with 5000 time steps each. The variables are injected with random linear cause-effect structures with time lags up to 200 steps using Tigramite \cite{Runge2023} to construct data with known ground truth. The randomly generated features may have vastly different numeric scales. Thus, we use min-max scaling per feature in each context window and report the Mean Average Percentage Error (MAPE). Furthermore, we identified the synchronized cause-effect pairs in a data-driven way (see Sec.~\ref{subsec:leadlagsync}) and created the training and test samples accordingly (see Sec.~\ref{subsec:dataaugmentation}). The train and test, i.e. cross-validation and evaluation, are performed in a sliding-window fashion, yielding a high amount of training and test samples. We use cross-validation to estimate the model's performance better and use 70\% train, 20\% test and 10\% validation samples, respectively. The models are trained for 500 epochs and a batch size of 1024 using one Nvidia A100 GPU. We highlight the performance improvement of using synchronized cause-effect pairs in comparison to non-synchronized data in the following Tab.~\ref{tab:performance_comparison}.

\begin{table}[ht]
    \centering
    \caption{Performance comparison of various state-of-the-art multivariate forecasting models trained on non-synchronized (NS) and synchronized (S) synthetic data. Mean Average Percentage Errors (MAPE) are reported, and the percentage difference in MAPE is used to highlight the performance improvements. Each model is fitted using cross-validation and subsequently evaluated to retrieve the MAPE.}
    \begin{tabular*}{\hsize}{@{\extracolsep{\fill}}lccc@{}}
        \hline
        \textbf{Model} & \textbf{MAPE (NS)} & \textbf{MAPE (S)} & \textbf{Diff. MAPE (\%)} \\
        \hline
        TSMixerx\cite{chen2023} & 7.24 & 5.07 & 29.97 \\
        MLPMultivariate\cite{fukushima1975} & 7.09 & 4.83 & 31.88 \\
        BiTCN\cite{sprangers2021} & 3.33 & 2.98 & 10.51 \\
        TFT\cite{lim2020} & 3.29 & 2.99 & 9.12 \\
        \hline
    \end{tabular*}
    \label{tab:performance_comparison}
\end{table}

The performance improvements due to synchronization of cause and effect are substantial. As we shift cause variables to synchronize with the effects and shift causes from outside the context window in case the cause happened sometime before the context, an improvement for all the models was expected. It is, however, essential to note that the improvements are heavily dependent on causal pair extraction and lag estimation, and we again highlight the fact that both causal graphs and lags need to be checked with domain experts in real-world environments. We can observe quite substantial MAPE deviations across the different models. This might be due to the fact that we train each model with the same configurations, i.e., the same scaler, the same number of epochs, etc., despite having vastly different amounts of learnable parameters. The relatively limited size of the datasets used in the experiment may also negatively affect some models' convergence.

\subsection{Experiment 2 - Transferability of Cause-Effect Pre-Training}

In this experiment, we evaluate the transferability of learned cause-effect dependencies by pre-training models on multiple synthetic source datasets and assessing their performance on a target dataset with unseen causal structures. Specifically, we investigate whether pre-training on a large and diverse set of causal relationships can improve forecasting accuracy on new datasets compared to models trained solely on limited target domain data. We compare two dataset configurations: models trained with non-synchronized cause-effect pairs and synchronized cause-effect pairs. We generated 100 synthetic source datasets, each consisting of a varying number of variables and 5000 time steps. As in Experiment 1, the variables are injected with random linear cause-effect structures using Tigramite. We compare source pre-trained and target fine-tuned models with models only trained on the target data. We average across the models and report the averaged MAPE of this experiment in Tab.~\ref{tab:unseen_performance}. We can observe that pre-training on many different causal structures improves accuracy over solely supervised training on the limited target data with unseen causal structures, indicating that the models generalize from seeing independent cause-effect samples to some extent. It is important to note that this experiment does not evaluate zero-shot capabilities. However, this is planned as part of future research. 

\begin{table}[h]
    \centering
    \caption{Performance comparison of pre-training + fine-tuning and supervised end-to-end training on target only applied to synchronized and non-synchronized cause-effect pairs. Mean Average Percentage Errors (MAPE) are reported for each scenario and are averaged across the models.}
    \begin{tabular*}{\hsize}{@{\extracolsep{\fill}}lcc@{}}
        \hline
        \textbf{Model Type} & \textbf{Dataset} & \textbf{MAPE} \\
        \hline
        Pre-Trained on Source, Fine-Tuned on Target & Synchronized & 2.21 \\
        Pre-Trained on Source, Fine-Tuned on Target & Non-Synchronized & 2.56 \\
        Supervised Training on Target Only & Synchronized & 2.38 \\
        Supervised Training on Target Only & Non-Synchronized & 2.42 \\
        \hline
    \end{tabular*}
    \label{tab:unseen_performance}
\end{table}

\section{Discussion and Conclusion}
\label{sec:conclusion}
This section discusses the stated research questions based on the research and experimental results conducted and highlights the assumptions for this methodology. In addition, we discuss the relevancy of (self-) supervised learning and transfer-learning for the process industry, their impact on Digital Twin modelling, and future research steps.

\subsection{Evaluating the Research Questions (RQs)}

\begin{enumerate}[label=\textbf{Research Question \arabic*:}, leftmargin=*]
    \item \textit{How can highly lagged, linear cause-effect relationships in multi-dimensional process data be identified in a data-driven way?}\\We utilize a data-driven methodology based on the Granger Causality to identify cause-effect pairs (see Ammann et al. \cite{Ammann2024}). We assume that the causal relationships can be linearly modelled using a VAR model. Non-linear Granger Causality methods exist; however, we follow a linear approach for simplicity. We highlight the necessity of non-linearity and the inclusion of expert knowledge in order to achieve high-quality causal discovery of complex systems.
    \item \textit{How can one synchronize cause-effect pairs w.r.t causal lag and create context-horizon training samples for CD pre-training without losing information in the sample?}\\We propose a target-oriented shift trick (see Fig.~\ref{fig:CauseEffectPair}), similarly to \cite{Zhao2024}. We recognize the complex causal structures, sometimes with isolated causal clusters in industrial multi-dimensional process data, making a correct synchronization across all $N$ variables a challenging task, as one might introduce wrong shifts due to complex transitive or isolated dependencies. Thus, we propose a novel data preparation method to break down the complex causal graph into pairs of cause and effect variables, mitigating unnecessary noise and complexity that might be introduced by CD methods training with out-of-sync or obsolete values in the samples. The proposed input-output format for the regression task is depicted in Fig.~\ref{fig:trainingpairs}. As we lose information on the right (and left, but this is filled up) side of the window for the cause variable due to shifting the latest values out of the window, we also include the non-synchronized cause variable in the input sample to prevent any loss of data. The proposed method is generally formulated and breaks down the problem of multi-dimensional time-series forecasting in complex causal structures into its core components, i.e. cause-effect pairs. The input-output format is constructed in such a way that essentially, we combine CD on the level of individual cause-effect pairs, assuming CI-like learning behaviour of the whole causal structure. 
    \item \textit{Do CD pre-training methods (e.g., TSMixerx, TFT, MLPMultivariate) benefit from synchronized cause-effect pairs for regression-related tasks like forecasting?}\\We evaluate the performance of various state-of-the-art forecasting models trained on non-synchronized (NS) and synchronized (S) synthetic data with known cause-effect relationships. We denote the percentage difference in MAPE to highlight the performance improvements. Each model is fitted using cross-validation and subsequently evaluated to retrieve the MAPE. We denote substantial percentage improvements across all models trained on the synchronized data compared to training on non-synchronized data. Details of all the experimental results are depicted in Tab.~\ref{tab:performance_comparison} and Tab.~\ref{tab:unseen_performance}. It is essential to highlight that the tests in this paper are conducted on synthetically generated datasets. We aim to thoroughly benchmark the proposed methodology (i.e., causal synchronization and training on cause-effect pairs) on real-world datasets. We will prioritize benchmarking with real-world data of industrial use-case partners, where we can validate the methodology using domain knowledge, i.e. domain experts knowing true causal relationships and the approximate inertia. We aim to pre-train several CD models across various industrial use-cases data using cause-effect pairs.
\end{enumerate}

Traditional models typically need significant recalibration to perform well in changing conditions, such as process state shifts or causal structure changes. However, by leveraging (self-) supervised learning and to extract representations from (un-) labeled data and utilizing transfer learning to adapt pre-trained models to diverse tasks, Digital Twins can achieve greater accuracy with reduced training. This allows DTs to efficiently adjust to the wide range of use cases expected in industrial applications, even when labelled data is limited. Training general-purpose time-series models with CD on synchronized cause-effect pairs aids in capturing causal cross-variable dependencies and enables the includance of highly lagged behaviours that may lie outside of the conditioned context window, an important aspect when modelling industrial processes with potential high inertia and slow-moving cause-effects. This research tackles several pain points of pre-training on multi-dimensional datasets with complex causal structures. 
It is important to emphasize that using data-driven methods to extract cause-effect relationships in real-world scenarios requires expert knowledge to verify or adjust the detected structures. Furthermore, the data-driven causal discovery in this research does not address non-linear cause-effect dependencies, a crucial aspect of complex systems. This further underscores the need for expert input in this methodology.
Furthermore, we currently model only pairs of interacting variables, meaning that the states of other variables in the broader causal structure are not considered during inference. We believe that incorporating information about the entire causal structure into the modelled cause-effect pair could significantly enhance our approach. Exploring reconciliation strategies, such as combining model results with the overall causal structure or adjusting the loss function with a regularization term to ensure stable predictions across different causal structures, is an area of high interest for our future research.

\subsection{Conclusion}
This paper proposes a novel pre-training strategy to address several critical challenges encountered when pre-training on diverse industrial datasets, particularly those arising from multi-dimensional time-series data, lagged cause-effect dependencies, complex causal structures, and varying numbers of variables.
Firstly, we utilize a data-driven methodology leveraging Granger Causality within a VAR model framework (see Ammann et al. \cite{Ammann2024}) to identify cause-effect pairs, utilizing a linear approach while recognizing the potential for future research in non-linear methods. Secondly, we introduce a synchronization technique called the target-oriented shift trick, similar to Zhao et al. \cite{Zhao2024}, however, we align individual cause-effect pairs w.r.t. causal lag, thereby creating context-horizon training samples (see Fig.~\ref{fig:trainingpairs}) for a novel pre-training strategy. This method tackles the complexity of industrial multidimensional process data by breaking the complex causal structure into manageable cause-effect pairs, minimizing the noise of misaligned time-series data, and ensuring more robust training in datasets with varying amounts of variables and causal structures. By utilizing the proposed pre-training strategy, we can train CD models on various cause-effect pairs across different multi-dimensional datasets, focusing on directly learning cause-effect behaviours on the variable pair level (CD) and indirectly learning from other cause-effects through shared weights (CI) (see Nie et al. \cite{Nie2023}) which is a novel contribution in this area of research. Our method specifically captures cause-effect behaviours, an essential part of simulation and control in industrial processes. \\
We demonstrate the performance benefits of cause-effect synchronization for regression tasks, such as forecasting using various state-of-the-art models on synthetic datasets. The results show significant improvements in model accuracy when trained on synchronized data compared to non-synchronized data (see Tab.~\ref{tab:performance_comparison}). In addition, we show that by pre-training on many different causal structures, we can improve accuracy over solely supervised training on the limited target data with unseen causal structures, indicating that the models generalize from seeing independent cause-effect samples (see Tab.~\ref{tab:unseen_performance}).\\
Future work will focus on zero-shot experiments, benchmarking the proposed methodology on real-world industrial datasets, validating the approach with domain-specific knowledge, and pre-training foundational CD models across diverse industrial use cases. We think this pre-training strategy may significantly impact diverse industrial use cases and enhance Digital Twin modelling. We believe that incorporating information about the entire causal structure into the modelled cause-effect pair could enhance our approach. Exploring reconciliation strategies, such as combining modelling results with the overall causal structure or adjusting the loss function with a regularization term to ensure stable predictions across different causal structures, is also an area of high interest for our future research.

\section*{Acknowledgments}
This work received funding as part of the Trineflex project (trineflex.eu), which has received funding from the European Union’s Horizon Europe research and innovation programme under Grant Agreement No 101058174. Funded by the European Union. In addition this work received funding from the Federal Ministry for Climate Action, Environment, Energy, Mobility, Innovation, and Technology (BMK), the Federal Ministry for Digital and Economic Affairs (BMDW), and the State of Upper Austria in the frame of SCCH, a center in the COMET - Competence Centers for Excellent Technologies Programme.



\bibliography{bib.bib}
\bibliographystyle{elsarticle-harv.bst}




\end{document}